\documentclass{egpubl}

\ifpdf \usepackage[pdftex]{graphicx} \pdfcompresslevel=9
\else \usepackage[dvips]{graphicx} \fi

\usepackage{mylib}

\SpecialIssueSubmission    
\usepackage[T1]{fontenc}
\usepackage{dfadobe}  

\usepackage{cite}  
\BibtexOrBiblatex

\electronicVersion
\PrintedOrElectronic

\ifpdf \usepackage[pdftex]{graphicx} \pdfcompresslevel=9
\else \usepackage[dvips]{graphicx} \fi

\usepackage{egweblnk} 

\title[Factored Neural Representation]{Factored Neural Representation for Scene Understanding}

\author[Y. Wong and N. J. Mitra]
{\parbox{\textwidth}{\centering 
        Yu-Shiang Wong$^{1}$\hspace{20pt} 
        Niloy J. Mitra$^{1,2}$
}\\
{\parbox{\textwidth}{\centering $^1$University College London \hspace{20pt} 
         $^2$Adobe Research 
        }
}
}

\usepackage{outlines}
\usepackage{soul} 
\usepackage[dvipsnames]{xcolor} 

\usepackage[T1]{fontenc}
\usepackage{dfadobe}  
\usepackage{amsfonts}
\usepackage{microtype}
\usepackage{overpic}
\usepackage{xspace}  

\usepackage{cite}  
\usepackage{amsmath}
\usepackage{xcolor}
\usepackage{stackengine}
\usepackage{multirow} 
\usepackage{array}

\usepackage{pifont}

\newcommand{\etal}{\textit{et al.}}
\newcommand{\para}[1]{\noindent {\textbf{#1}}}

\newcommand{\ray}{{\mathbf{r}}}

\newcommand{\rayColor}{{C}(\ray)} 
\newcommand{\rayDepthGT}{D(\ray)}

\newcommand{\pIdx}{i}
\newcommand{\pColor}{{\mathbf{c}}}   
\newcommand{\quadstep}{\delta}
 
\newcommand{\pDensity}{\sigma} 
\newcommand{\pTransmit}{T} 
\newcommand{\pSDF}{\psi} 

\newcommand{\sample}{\mathbf{p}}

\newcommand{\freeSamples}{P_{\mathrm{free}}}

\newcommand{\camera}{\Pi}
\newcommand{\render}{\mathcal{R}}

\newcommand{\facBkg}{f^0}
\newcommand{\facObj}[1]{f^{#1}}
\newcommand{\bbox}[1]{B_{#1}}
\newcommand{\transform}[1]{\mathbf{T}_{#1}}
\newcommand{\facRep}{\mathcal{F}}

\newcommand{\mycaption}[2]{\caption{{\textbf{#1.}} {#2}}}

\newcommand{\imap}{\textsc{iMAP}\xspace}
\newcommand{\nice}{\textsc{NiceSLAM}\xspace}
\newcommand{\behave}{\textsc{Behave}\xspace} 
\newcommand{\scene}{\textsc{SYN-Scene}\xspace}

\newcommand{\pOpacity}{\alpha} 
\newcommand{\pixelxy}{uv} 
\newcommand\xrowht[2][0]{\addstackgap[.5\dimexpr#2\relax]{\vphantom{#1}}}

\newcolumntype{C}[1]{>{\centering}m{#1}}

\newcommand{\vv}{\ding{51}}%

\begin{document}
\teaser{
 \includegraphics[width=\linewidth]{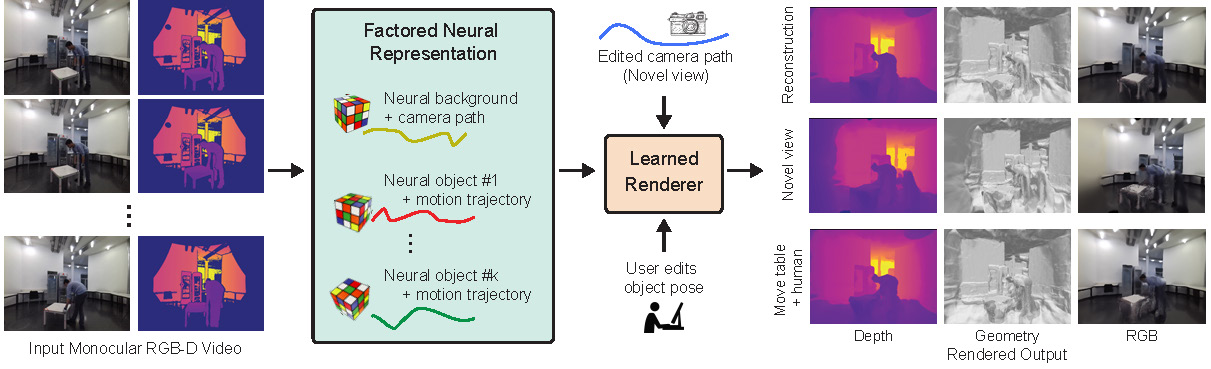}
 \centering
  \caption{
  We present an algorithm that directly factorizes raw RGB-D monocular video~(sequence from \cite{bhatnagar2022behave} in this example) to produce object-level neural representation with motion trajectories and nonrigid deformation information. The decoupling subsequently enables different object manipulation and novel view synthesis applications to produce authored videos. We do not use object templates or motion prior but instead use an end-to-end optimization to enable the factorization. Note that the human subject deforms and moves in this sequence. 
  }
\label{fig:teaser}
}
\maketitle
\begin{abstract}
A long-standing goal in scene understanding is to obtain interpretable and editable representations that can be directly constructed from a raw monocular RGB-D video, without requiring specialized hardware setup or priors. The problem is significantly more challenging in the presence of multiple moving and/or deforming objects. 
Traditional methods have approached the setup with a mix of simplifications, scene priors,  pretrained templates, or known deformation models. 
The advent of neural representations, especially neural implicit representations and radiance fields, opens the possibility of end-to-end optimization to collectively capture geometry,  appearance, and object motion. However, current approaches produce global scene encoding, assume multiview capture with limited or no motion in the scenes, and do not facilitate easy manipulation beyond novel view synthesis. 
In this work, we introduce a factored neural scene representation that can directly be learned from a monocular RGB-D video to produce object-level neural presentations with an explicit encoding of object movement (e.g., rigid trajectory) and/or deformations (e.g., nonrigid movement). 
We evaluate ours against a set of neural approaches on both synthetic and real data to demonstrate that the representation is efficient, interpretable, and editable (e.g., change object trajectory). 
Code and data are available at: \href{http://geometry.cs.ucl.ac.uk/projects/2023/factorednerf/}{http://geometry.cs.ucl.ac.uk/projects/2023/factorednerf/}.

\end{abstract}  

\section{Introduction}

\textit{Scene understanding from video capture} has a long history in content creation. It subsequently enables editing by replaying the content from novel viewpoints and allowing object-level modifications. The task is particularly challenging in the dynamic context of moving and deforming objects when observed through a moving (monocular) camera. Traditional approaches make simplifications by assuming the scene to be static~\cite{curless1996volumetric}, or requiring access to a variety of priors in the form of object templates~\cite{choi2013rgbd,ren2013star3d}, deformable object models~\cite{blanz1999morphable,anguelov2005scape,loper2015smpl}, or simultaneous localization and mapping~\cite{newcombe2011dtam,izadi2011kinectfusion}. Additional complexity arising  from unknown objects' appearance is ignored. 

Neural Radiance Field~(NeRF)~\cite{mildenhall2020nerf}, a new volumetric neural representation,  provided a breakthrough in terms of producing highly photorealistic (static) representation, simultaneously capturing  geometry and appearance from only a set of posed images. A substantial body of work has rapidly emerged to extend the formulation to dynamic settings~\cite{li2022neural,du2021neural,pumarola2021d,li2021neural,tretschk2021non,xian2021space,gao2021dynamic}, work with localized representations for real-time inference~\cite{liu2020neural,reiser2021kilonerf,yu2021plenoctrees,lombardi2021mixture,sun2022direct,karnewar2022relu,fang2022fast,wang2022fourier}, support fast training~\cite{deng2022depth,sun2022direct,karnewar2022relu,fang2022fast,liu2022devrf}, and investigate applications in the context of generative models~\cite{karnewar_3InGan_3dv_22}.  However, such representations often lack interpretability, require multiview input, fail to provide scene understanding, 
and do not provide object-level factorization or enable object-level scene manipulation. 

We introduce \textit{factored neural representation}. This object-level scene representation supports interpretability and editability while capturing geometric and appearance details under object movement and viewpoint changes. Our approach does not require any object template, deformation prior, or pretraining object NeRFs. Starting from an RGB-D monocular video of a dynamic scene, we demonstrate how such a factored neural representation can be robustly extracted via joint optimization by leveraging off-the-shelf image-space segmentation and tracking information. Factorization is provided through object-level neural representations and object trajectory and/or deformations. 

Technically, we formulate a global optimization to simultaneously build and track per-object neural representations along with a background model while solving for object trajectories and camera path. Further, we model deformable bodies (e.g., a moving human) by adapting the learned neural representation over time. Our proposed representation combines the advantages of object-centric representations and motion tracking, thereby allowing per object manipulation, without having to pay the overhead of separately building object priors or requiring 3D supervision, and naturally integrates information from a monocular input over time across the neural representations to recover from occlusion. For example, Figure~\ref{fig:teaser} shows a factored representation obtained by our method by operating on a monocular RGB-D sequence~\cite{bhatnagar2022behave} of $60$ frames, along with some edits. 

We evaluate on both synthetic and real scenes. We compare to competing methods and show that ours can produce better object representations and camera/object trajectories. 
Note that prior methods often focus on 
only rigid motion and separated optimization~\cite{wong2021rigidfusion,mueller2021completetracking},
assume access to geometric priors~\cite{mueller2021completetracking},
a single non-rigid object with local motion~\cite{park2021hypernerf, cai2022neural}, 
a global representation~\cite{li2021neural,tretschk2021non,xian2021space},
foreground-background separation and novel-view rendering without geometric  reconstruction~\cite{gao2021dynamic,yuan2021star,wu2022d,song2023nerfplayer},
or static scenes with an implicit representation~\cite{zhu2022nice,sucar2021imap,yu2022monosdf}. 
We relax many of these restrictions and  demonstrate that our factorized representation naturally enables edits involving object-level manipulations. In summary, we introduce a \textit{neural factored scene representation} and develop an end-to-end algorithm involving a joint optimization formulation to factorize monocular RGB-D videos directly. 

\section{Related Work}

\para{Scene reconstruction using traditional methods.}
Aggregating raw scans while simultaneously estimating and accounting for underlying camera motion is an established way of acquiring large-scale geometry of rigid scenes (e.g., KinectFusion~\cite{izadi2011kinectfusion}, VoxelHash~\cite{niessner2013real}). 
This paradigm has been extended for dynamic scenes by simultaneously segmenting and tracking multiple (rigid) objects (e.g., CoFusion~\cite{runz2017co}, MaskFusion~\cite{runz2018maskfusion}, MidFusion~\cite{xu2019mid}, EmFusion~\cite{strecke2019fusion}, RigidFusion~\cite{wong2021rigidfusion}) or, decoupling the handling of objects and human motion (e.g., MixedFusion~\cite{zhang2017mixedfusion}). 
These methods explicitly track and represent geometry, without or with textured colors, do not support joint optimization, and need special handling for multiple objects.

\para{Neural implicit representation.}
In the context of object representation, 
the recent introduction of the neural implicit representation~\cite{park2019deepsdf,mescheder2019occupancy,chen2019learning} has resulted in an explosion of works to overfit a single object or to encode object collections. 
Researchers have proposed improvements to better capture high-frequency details~\cite{tancik2020fourier,mehta2021modulated,sitzmann2020implicit}, and investigated hybrid implicit representations like point-based~\cite{erler2020points2surf,cai2020learning,li2022learning,zhang20223dilg}, surface-based~\cite{ genova2020local,chabra2020deep,morreale2022neural}, or grid-based methods~\cite{takikawa2021neural,chibane2020implicit,peng2020convolutional,karnewar2022relu}
to achieve better trade-offs among inference speed, memory footprint, and locality of representations. These works couple geometry and appearance captures but largely focus on static, individual objects and do not model changing (object) configurations.

\para{Neural representations through image guidance.} 
In the context of joint material and geometry representation, differential rendering directly optimizes neural implicit representations using only RGB images for supervision. This is achieved by either ray tracing or volumetric rendering based approaches. %
Ray tracing accounts for explicit surface intersection and calculates gradient on the surface using implicit differentiation~\cite{niemeyer2020diff, yariv2020multiview}, max pooling~\cite{liu2019learning}, or unfolding sphere tracing~\cite{kellnhofer2021neural,liu2020dist,jiang2020sdfdiff}.
However, for objects with complex topologies, these methods suffer from hard-to-propagate local gradients. 
In contrast, volumetric rendering~\cite{max1995optical,henzler2019platonicgan}, leading to Neural Radiance Fields~(NeRF)~\cite{mildenhall2020nerf}, integrates density and color samples along rays by modeling a radiance field and employs a coarse-to-fine sampling scheme to focus on surface density, without explicitly distilling the underlying geometry. When converted from the density field, the learned implicit geometry is usually noisy and inaccurate. Again these approaches focus on isolated objects.

\para{Neural scene representation.}
In scene analysis, combining volumetric rendering with an implicit representation~\cite{mildenhall2020nerf,wang2021neus,oechsle2021unisurf} has led to a series of works revisiting traditional scene representations. For example, methods have been proposed for the 3D reconstruction and scene editing tasks, including indoor scene reconstruction~\cite{azinovic2022neural,yu2022monosdf}, structure from motion~\cite{martin2021nerf}, simultaneous localization and mapping~\cite{sucar2021imap,zhu2022nice}, bundle adjustment~\cite{azinovic2022neural,clark2022volumetric,lin2021barf}, multi-view stereo~\cite{wei2021nerfingmvs}, scene reconstruction using ellipsoid proxies~\cite{Zhang_2023_CVPR}, surface meshing and interactive editing~\cite{garbin2022voltemorph,jambon2023nerfshop}, distilling segmentation priors to extract instances using a single neural network~\cite{kobayashi2022distilledfeaturefields,wang2023dmnerf}. 
Most of these works, however, focus on static scenes. In Section~\ref{sec:results}, we present several comparisons with \imap~\cite{sucar2021imap} and \nice~\cite{zhu2022nice} that perform implicit scene representation with simultaneous tracking but focus on global scene representations with static objects. 

\para{Modeling dynamic objects.} 
In order to obtain NeRF representations for deforming objects, parametric and non-parametric template models have been exploited. When parametric models are available (e.g., for human bodies), the underlying parametric template is utilized to create a part-based NeRF representation~\cite{peng2021neural,chen2021animatable,gafni2021dynamic,liu2021neural,noguchi2021neural,weng2022humannerf}, i.e., each part having a corresponding NeRF encoding, to create dynamic avatar models with pose and shape control. 
For non-parametric models, dynamic NeRF has been proposed by solving for a template representation and capturing dynamic appearance by reindexing into a base (i.e., canonical model) NeRF representation~\cite{pumarola2021d,xu2021h,park2021nerfies,park2021hypernerf,fang2022fast,wang2022fourier,cai2022neural,li2021neural,gao2021dynamic}. 
Such representations are then used to model rigidly moving objects assuming access to static pretrained NeRF~\cite{yen2021inerf}, predict object-space normalized coordinates for 6 DOF extraction and tracking~\cite{li2022nerf}, perform point-to-SDF tracking~\cite{ueda2022neural}, or predict surface correspondences~\cite{huang2022neural}. These methods focus on objects in isolation. 

\para{Modeling dynamic scenes.} 
Recent works have trained global object NeRF from monocular input~\cite{li2021neural,gao2021dynamic}, capture dynamic effects by overfitting to a global 4D space-time volume~\cite{xian2021space,cao2023hexplane,fridovich2023k}, 
and explicitly capture human interactions~\cite{jiang2022neuralhofusion,shuai2022novel}.
Researchers have investigated the effect of segmentation, tracking, and NeRF modeling tasks in other efforts.  
Notable examples include monocular with foreground and background decomposition~\cite{martin2021nerf,yuan2021star,wu2022d,song2023nerfplayer}, 
modeling rigid objects with a planar background model~\cite{ost2021neural,kundu2022panoptic},
egocentric video segmentation~\cite{tschernezki2021neuraldiff}, 
neural fusion fields~\cite{tschernezki2022neural}. 
We also develop a dynamic NeRF representation that can be extracted, without requiring pre-training and parametric templates, 
simultaneously with reconstruction for each object instead of only supporting novel-view rendering with a single dynamic element.
Further, we aggregate information across views to recover from occlusion. 
Once trained, our factored representation can be viewed from novel camera paths and used to make changes to object trajectories and placements.

\section{Image Formation Model}
\label{sec:imageFormation}

Before introducing the optimization formulation in Section~\ref{sec:algorithm}, we  present our image formation model to produce a rendered image from a factored neural representation. 

\para{Volume rendering.} 
To render an image $I$ from a given camera setup $\camera$, volume rendering~\cite{max1995optical,mildenhall2020nerf} 
maps each image pixel to form a camera ray $\ray$. Points are sampled on each such ray and sorted based on their depth values to produce a  rendered color $\rayColor$ as the integration of the sampled point colors $\{ \mathbf{ \pColor_{\pIdx} } \}$ weighted by the corresponding point density $\{ \pDensity_{\pIdx} \}$ and (accumulated) transmittance $\{ \pTransmit_{\pIdx} \}$. 
Note that samples along a ray $\ray:=(\mathbf{o},\mathbf{d})$, going through point $\mathbf{o}$ along a unit direction $\mathbf{d}$, are parameterized as $\sample(s_i) := \mathbf{o} + s_i \mathbf{d}$ for increasing scalar depth samples $s_i \in \mathbb{R}^+$. 
Using the samples, we discretize the continuous formulation using the quadrature approximation as:
\begin{eqnarray}\label{eq:volrender}
  \rayColor &:=& \sum_{\pIdx}^{} 
        \pTransmit_{\pIdx} \pOpacity_{\pIdx} \pColor_{\pIdx} \; \nonumber\\ 
  \pOpacity_{\pIdx} &:=& 1-\mathrm{exp}(-\pDensity_{\pIdx}\quadstep_{\pIdx})\nonumber\\
  \pTransmit_{\pIdx} &:=& 
        \prod_{j}^{i-1} (1 - \pOpacity_{\pIdx}) 
\end{eqnarray}
where $\quadstep_{\pIdx}$ is the depth distance between two adjacent samples and $\sigma_{\pIdx}$ is the predicted point density.
Recall that point opacity $\pOpacity_{\pIdx}$ represents the opacity of the point position $\sample(s_i)$, while the transmittance $ \pTransmit_{\pIdx} $ indicates the cumulative transmittance before a ray hits the $\pIdx$-th sample point. 
Looping over all the image pixels $\{\pixelxy\}$, we obtain 
$I := \render(\camera, f_{\theta}, \{\ray_{\pixelxy}\})$, where the function 
$f_\theta$, typically modeled by an MLP~\cite{mildenhall2020nerf}, can be probed to produce density and color samples as 
$f_\theta(\sample(s_i), \mathbf{d}):= (\sigma_\pIdx, \pColor_{\pIdx})$. Typically, only the color values are view dependent.  

\begin{figure}[b!]
    \centering
    \begin{overpic}[width=\columnwidth]{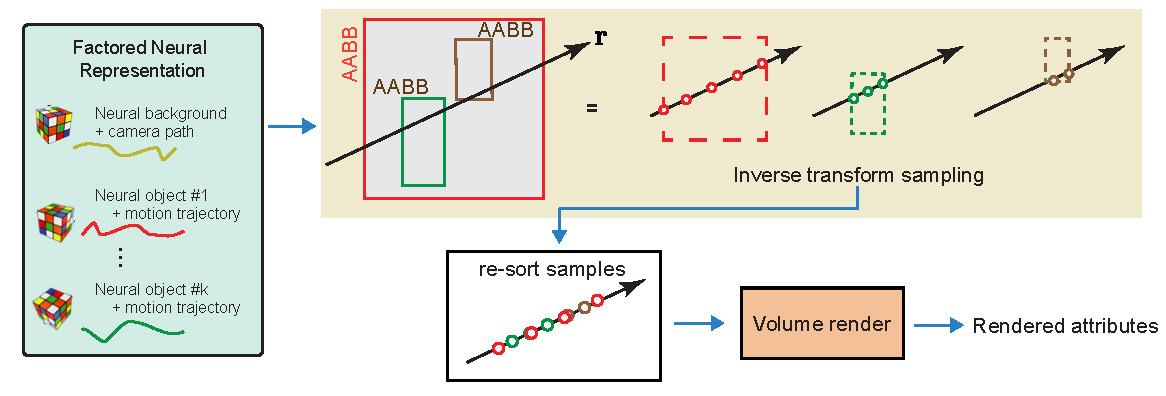}
    \end{overpic}
    \mycaption{Rendering neural factored representation}{Given a factored representation $\facRep\{ (\facObj{i}, \pSDF^i, \bbox{i}, \transform{i})_{i=0}^k \}$ and any query ray $\ray$ from the current camera, we first intersect each objects' bounding box $\bbox{i}$ to obtain a sampling range and then compute a uniform sampling for each of the intervals. For each such sample $\sample$, we lookup feature attributes by re-indexing using local coordinate $\transform{i}^{-1}\sample$, resort the samples across the different objects based on (sample) depth values, and then volume render to get a rendered attribute. Background is modeled as the $0$-th object. See Section~\ref{sec:imageFormation} for details. For objects with active \textit{nonrigid} flag, we also invoke the corresponding deformation block (see Section \ref{sec:algorithm} and Figure~\ref{fig:arch}). 
    The neural representations and the volume rendering functions are jointly trained. 
    }
    \label{fig:ray_illustration}
\end{figure}

\begin{figure*}[t!]
    \centering
    \begin{overpic}[width=\textwidth]{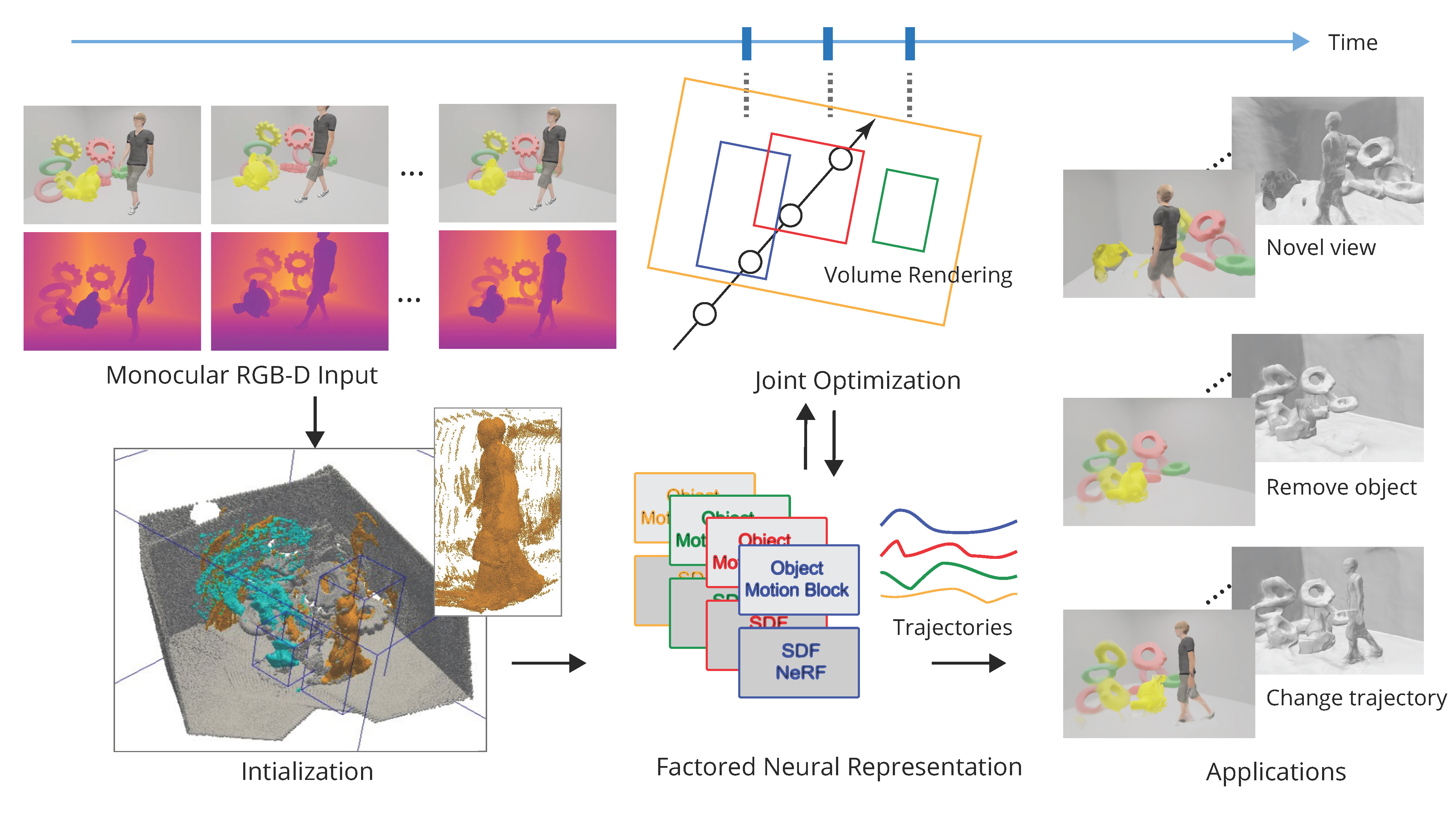}
        \put(26.5,29.8){$\{ I(t):= (C_t, D_t) \}$}
        \put(52,32){$A(\ray) := \render(\camera, \facRep,\ray)$ }
        \put(55,21){$
            \facRep := \{ (\facObj{i}, \pSDF^i, \bbox{i}, \transform{i})_{i=0}^k \}$
    }
\end{overpic}
    \mycaption{Method overview}
    {Starting from a monocular RGB-D sequence $\{I(t)\}$, we extract a \textbf{factored neural representation} $\facRep$ that contains separate neural models for the background and each of the moving objects along with their trajectories. For any object tagged as \texttt{nonrigid}, we also optimize a corresponding deformation block (e.g., human). 
    First, in an initialization phase, we assume access to keyframe annotation (segmentation and AABBs) over time, propagate the annotation to neighboring frames via dense visual tracking and optical flow, and estimate object trajectories.
    Then, we propose a joint optimization formulation to perform end-to-end optimization using a customized neural volume rendering block. The factored representation enables various applications involving novel view synthesis and object manipulations. 
    Please refer to the supplemental  showing  reconstruction quality and applications. }
    \label{fig:pipeline}
\end{figure*}

\para{Volume rendering with implicit surface.} 
Implicit surface representation, such as occupancy or signed distance fields, can also be used with volume rendering~\cite{wang2021neus,oechsle2021unisurf,yariv2021volume} and provides an inductive bias for modeling surface geometry. We found this more suitable for object-level factored representation as we can easily regularize the optimization to encode object surfaces instead of producing volumetric clouds. Here, we employ the signed distance field formulation proposed by Wang~\etal~\cite{wang2021neus} and convert the signed distance value $\pSDF$ to the density values by assigning non-zero values near the zero level set of the modeled surface geometry:  
\begin{equation}
\label{eqn:implicit2density}
\pOpacity_{j} \leftarrow max \left ( 
        \frac{ \Phi(\pSDF_j) -\Phi( \pSDF_{j+1} )}{ \Phi( \pSDF_j) },
              0 \right), 
\end{equation}
where we use a shorthand $\pSDF_j :=  \pSDF ( \sample(s_j))  $ for the $j$-th sample 
and 
$\Phi$ is the sigmoid function. 
Here, we represent the rendering function as 
$I := \render(\camera, f_{\theta}, \pSDF, \{\ray_{\pixelxy}\})$, where the function 
$f_\theta$ again can be probed to produce only view-dependent color samples $f_\theta(\sample(s_i),\camera):= \pColor_{\pIdx}$ and $\pSDF$ represents the learned SDF function. 

\para{Attributes rendering.} By replacing point color $\pColor_{\pIdx}$ with any other attribute $a_\pIdx$, such as depth~\cite{xian2021space,zhu2022nice} or semantic labels~\cite{zhi2021ilabel}, volumetric rendering can be generalized to render depth or semantic segmentation, respectively. 
Specifically, for any attribute $a_\pIdx$ and ray $\ray$, we simply compute an attribute as 
$A({\ray}) := \sum_{\pIdx}^{} \pTransmit_{\pIdx} \pOpacity_{\pIdx}  a_{\pIdx}$.

\para{Volume rendering with factored neural representation.} 
Our proposed factored representation $\facRep := \{ (\facObj{i}, \pSDF^i, \bbox{i}, \transform{i})_{i=0}^k \}$ for a background model $\facBkg$ and the foreground objects $\{\facObj{i}, i \in [1,k]\}$, which can be probed to output density and color attributes. Each model, the background or any foreground object, can be probed to output color attributes with corresponding AABB~(axis aligned bounding boxes) $\{ \bbox{i} \}$, transformations $\{ \transform{i} \}$ to map the AABB local coordinates to the global coordinate system, and implicit SDF functions $\{ \pSDF^i \}$ to produce density samples.  
We now define the rendering function 
$I := \render(\camera, \facRep, \{\ray_{\pixelxy}\})$ using our factored representation $\facRep$, with 
background and superscripts $i\in [1,k]$ denoting the $k$ foreground objects. 
Figure~\ref{fig:ray_illustration} illustrates the process. 
For each ray $\ray$, for each intersected model, computed using its AABB $\bbox{i}$, 
we obtain SDF density values using uniform samples and perform inverse transform sampling to generate 128 samples per ray. 
For the background ($i=0$) or foreground ($i\in [1,k])$ samples, we obtain 
$\facObj{i}(\transform{i}^{-1}\sample^i(s_j),\camera):= \pColor^i_j$ and density $\sigma^i_j$ using Equation~\ref{eqn:implicit2density} using $\pSDF^i$ using the remapped samples $\transform{i}^{-1}\sample^i(s_j)$, expressed in the local coordinate systems of the objects. We collect the samples across the background and all the intersecting objects, sort the samples based on their depth values, and volume render the colors/attributes as described earlier (see Equation~\ref{eq:volrender}). 

\section{Algorithm}
\label{sec:algorithm}

As input, we take in RGB-D frames, denoted by
$\{ I(t):= (C_t, D_t) \}$ with color $C_t$ and depth $D_t$ frames at time $t$,
of scenes with one or more moving objects, where objects can be moving rigidly or non-rigidly (e.g., humans). 
We assume access to keyframe annotation over time, containing instance segmentation, axis-aligned bounding boxes (AABBs), and \texttt{rigid} or \texttt{nonrigid} flags. In an initialization step, this information is used to extract initial camera and object trajectories and instance masks over time in the camera space. 
As output, we produce a factored neural representation $\facRep$ of the scene, where for each object we produce a neural representation along with its estimated object trajectory, and for a nonrigid object also an associated deformation function. 
In Section~\ref{sec:results}, we use these inferred factored representations to directly render novel view synthesis or perform object-level manipulations. 

To obtain such a factored representation, we have to address several challenges. 
First, the extracted segmentation information from the RGB-D frames is imperfect; hence, any information or supervision (e.g., segmentation loss) derived from them leads to error accumulation. 
Second, we must recover from artifacts in initial pose estimation, especially in scenes with insufficient textures to guide the camera calibration stage. Auto-focus, color correction, and error accumulation in real captures pose further challenges. 
Third, since we only use monocular input, the input provides partial information in the presence of occlusion, both in shape and appearance. Without priors, we have to recover from the missing information by fusing information across the (available) frames. 
Finally, we allow objects to exhibit nonrigid motion (e.g., human walking) and have to factorize object deformation from object motion. %
In the following, we present how to set up a joint optimization, with suitable initialization and regularizers, involving object tracking, neural representations, and volume rendering   to solve these challenges. 

\para{Initialization.}
We use an off-the-shelf visual tracker~\cite{wang2019fast} with keyframe annotation, including instance segmentation and AABBs, to propagate the keyframe segmentation across the frames. 
To get an initial registration, we run an optical flow network~\cite{teed2020raft} to find initial correspondences and solve for frame-to-frame rigid alignment using  iterated closest point~(ICP) approach. 
The registration information across frames provides object trajectory $\{ \transform{i}(t) \}$ estimates. 

\para{Joint optimization.}  
We now introduce the main loss terms to capture reconstruction quality and additional regularizers to get a desired factored representation. 

\noindent\textit{Reconstruction loss:} We render color and depth images using current (multi-object) neural factored representation as described in Section~\ref{sec:imageFormation}. Note that the object trajectories $\transform{i}$ are indexed by frame times, i.e., $\transform{i}(t)$. 
We compare the sampled color $C$ and depth $D$ attributes in a set of minibatch samples $P$ against the estimated attributes using the L1 reconstruction loss, i.e., 
\begin{align}
& L_\mathrm{color} (\facRep) := 
\sum_{ (\ray,t) \in P } \left \| C_t(\ray) - R_C(\camera, \facRep(t), \ray ) \right \| /{|P|} \ \:\text{and}
\nonumber \\
& L_\mathrm{depth} (\facRep) := 
\sum_{ (\ray,t) \in P }| D_t(\ray) - R_D(\camera, \facRep(t), \ray ) |/{|P|}.
\end{align} 
We render the current background and foreground neural objects to produce RGB and depth attributes and sum them up over the individual frames. 

\noindent\textit{Free-space loss:} 
One approach to check the factorization quality is to compare the predicted object segmentation, computed using the current re-projection of objects' transmission, against the input segmentation. However, this approach leads to poor results as segmentation estimates are noisy. Instead, we focus on the complement space and define a free-space loss~(cf., \cite{xian2021space}) to penalize density values in regions indicated to be free according to the raw depth information. 
For any point sampled from any of the objects, we want identify free-space samples using depth $\rayDepthGT$. Specifically, we constrain the integrated weights of each free-space sample $\sample \in \freeSamples$, before reaching the object point (i.e., zero-isosurface of $\pSDF^i$), to be zero using L1 loss. We found this loss to be better than a cross-entropy segmentation loss in the joint-training setting. Specifically, 
\begin{align}
    L_{\mathrm{free}}(\facRep) :=& 
                \sum_{\sample \in  \freeSamples} 
                    \left | \: 
                         \pTransmit_{\sample} \pOpacity_{\sample} 
                    \: \right | / {| \freeSamples|}  \nonumber \\
                    &  \textrm{where} \quad \freeSamples = 
                        \left \{ \sample(s,\ray)| s < D_t(\ray) \right \}.
\end{align}
\noindent\textit{Non-rigid deformation:} 
In order to handle non-rigid objects, we additionally incorporate a deformation block, for objects marked with flag \texttt{nonrigid}. Specifically, we adopt a state-of-the-art bijective deformation network proposed by Cai~\etal~\cite{cai2022neural}, which consists of three sub-networks, each predicting a low-dimensional deformation. Given an input 3D point, each sub-network selects one axis, predicts a 1D displacement, and infers a 2D translation and rotation for the other axes. These sub-networks are sequentially invoked in the XYZ axis order. Note that this block gets directly optimized via the reconstruction loss and is \textit{not} supervised with ground truth deformation. 

\noindent\textit{Surface regularizers:} 
In order to regularize our network to output a canonical model, we employ auxiliary losses to constrain our geometry models to be actual surfaces by penalizing the implicit functions $\pSDF^i$ to (i)~be a true signed distance field (i.e., using Eikonal loss) ; (ii)~requiring the surface points (i.e., points within $\pm \epsilon$ of the zero level set of the SDFs denoted by $\Omega_\epsilon({\pSDF^i})$) to have normals in the direction of normals $\mathbf{n}(\mathbf{x})$ estimated from the input RGB-D~\cite{GuerreroEtAl:PCPNet:EG:2018}; and 
(iii)~surface points to have zero implicit values. 
These auxiliary losses does not slow down the optimization since they can be directly calculated without performing volumetric rendering.
Putting them together we get, 
\begin{align} 
  L_{\mathrm{surface}}(\facRep) :=
   \frac{1}{(k+1)}\sum_{i \in [0,k] }  & \bigg[ 
     \sum_{\mathbf{x} \in P_{\bbox{i}}} |  \| \nabla \pSDF^i(\mathbf{x})  \|_2 - 1 | / |P_{\bbox{i}}|
     \nonumber \\ 
  & + 
  \sum_{\mathbf{x} \in P_{\Omega_i}} |1- < \nabla \pSDF^i(\mathbf{x}), \mathbf{n}(\mathbf{x})>|  / |P_{\Omega_i}|
  \nonumber \\ 
  & + 
  \sum_{\mathbf{x} \in P_{\Omega_i}} |\pSDF^i| / |P_{\Omega_i}| \bigg] ,
\end{align}
where $P_{\bbox{i}}$ and $P_{\Omega_i}$ denote the randomly sampled spatial points and surface samples in the object bounding box $\bbox{i}$, respectively. 
Finally, we arrive at the full optimization problem as, 
\begin{align}
    \min_{\facRep}
    L_{\mathrm{total}}(\facRep) := \; & L_\mathrm{color} (\facRep) + \lambda_1 L_\mathrm{depth}(\facRep) \nonumber \\
     &+ \lambda_2 L_{\mathrm{free}}(\facRep) 
     + \lambda_3 L_{\mathrm{surface}}(\facRep),
\end{align}
We use $\lambda_1=0.1$, $\lambda_2=1.0$, and $\lambda_3=0.1$ in our experiments where $\lambda_1 < 1$ due to noisy depth input. 
Recall that the factored representation $\facRep := \{ (\facObj{i}, \pSDF^i, \bbox{i}, \transform{i})_{i=0}^k \}$ maintains a specialized model for the background and each of the object trajectories $\transform{i}(t)$ being time dependent. 

\section{Evaluation}
\label{sec:results}

\begin{figure}[b!]
    \centering
    \includegraphics[width=\columnwidth]{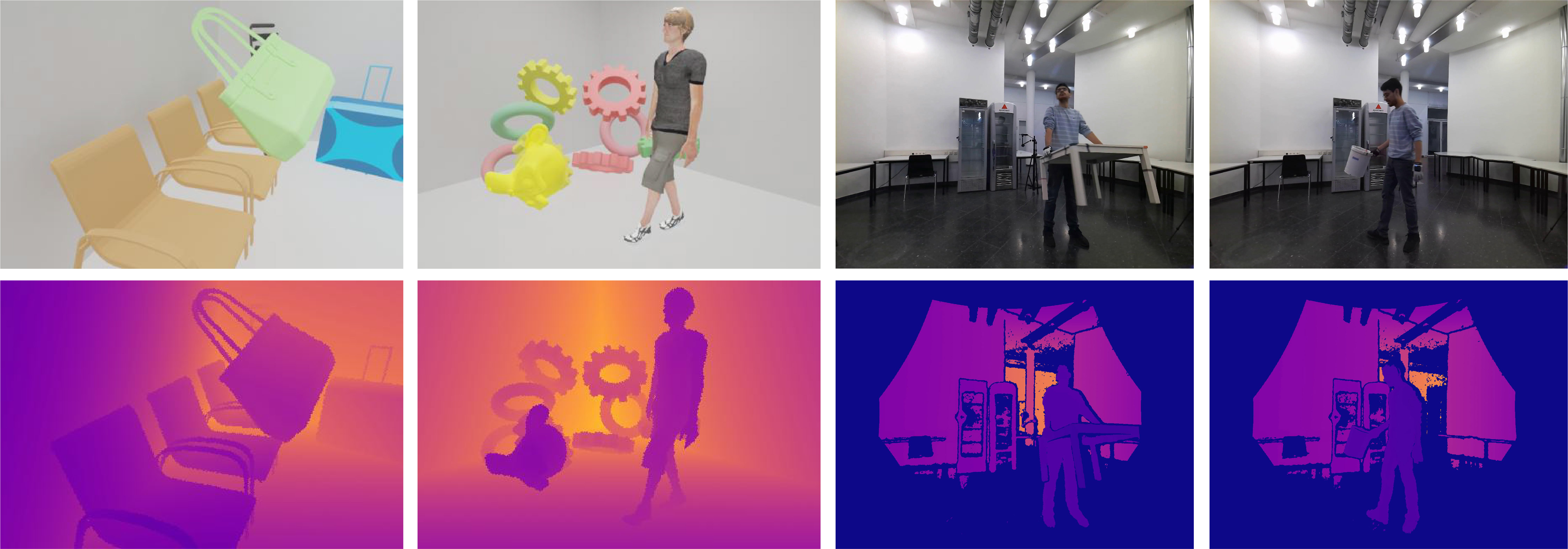}
    \mycaption{Dataset}{
    We test on a mix of dynamic datasets, including synthetic scans (\scene A, B, and C) and real RGB-D monocular captures from \behave~\cite{bhatnagar2022behave} (tablesquare-move, trashbin, yogaball-play, chairblack-lift). Here we show
    some representative frames, 
    RGB~(top) and depth~(bottom).
    }
    \label{fig:dataset}
\end{figure}

We evaluated Factored Neural Representations on a variety of synthetic and real scenes, in the presence of rigid and nonrigid objects. 
In each case, we start with only RGB-D sequences, without access to any object template.  

\para{Dataset.}
We tested on two types of datasets, synthetic and real. 
As \textit{synthetic dataset}, we propose a new dataset using public available CAD models~\cite{chang2015shapenet,greff2022kubric} and render RGB-D sequences using Blender~\cite{greff2022kubric,blender} with simulated sensor noises~\cite{handa2014benchmark}. To inject motion, we manually edit camera motion, rigid object motion, and combine non-rigid motion from the DeformingThings4D~\cite{li20214dcomplete} dataset.  
As representative examples, we present three sequences, \scene A, B, and C, each spanning for 90-100 frames and containing multiple dynamic objects. 
For these synthetic sequences, we have access to ground truth data (e.g., object trajectory, object segmentation, deformation model). 
\textit{This new dataset will be made publicly available on publication}.

As \textit{real dataset}, we use the \behave~\cite{bhatnagar2022behave} dataset, which provides human object interaction RGB-D videos with keyframe annotation. We crop and evaluate the first non-occlusion sequence in each scene to avoid the object re-identification issue.
Figure~\ref{fig:dataset} shows some representative frames. 

\begin{table*}[!]
\centering
\mycaption{Reconstruction error on our synthetic dataset}{
(Top/Bottom)~Quantitative color/depth novel view rendering results on validation cameras. 
Ours largely produces better reconstruction, validating that our joint optimization captures better scene geometry.
%
}
\label{tab_eval_syn_novel}

\resizebox{\textwidth}{!}{%

\begin{tabular}{|cccccccccccc|}
\hline
\multicolumn{12}{|c|}{\begin{tabular}[c]{@{}c@{}}Color Reconstruction (PSNR $\uparrow$ / SSIM $\uparrow$) \xrowht{15pt}\end{tabular}} \\ \hline 
\multicolumn{1}{|c|}{} &
  \multicolumn{3}{c|}{SYN-Scene-A} &
  \multicolumn{4}{c|}{SYN-Scene-B} &
  \multicolumn{4}{c|}{SYN-Scene-C} \\ \hline
\multicolumn{1}{|c|}{} &
  \multicolumn{1}{c|}{Full} &
  \multicolumn{1}{c|}{BG} &
  \multicolumn{1}{c|}{FG1} &
  \multicolumn{1}{c|}{Full} &
  \multicolumn{1}{c|}{BG} &
  \multicolumn{1}{c|}{FG1} &
  \multicolumn{1}{c|}{FG2} &
  \multicolumn{1}{c|}{Full} &
  \multicolumn{1}{c|}{BG} &
  \multicolumn{1}{c|}{FG1} &
  FG2 \\ \hline
\multicolumn{1}{|l|}{\imap} &
  \multicolumn{1}{l|}{20.32/0.79} &
  \multicolumn{1}{l|}{21.02/0.86} &
  \multicolumn{1}{l|}{15.02/0.88} &
  \multicolumn{1}{l|}{17.98/0.78} &
  \multicolumn{1}{l|}{21.87/0.90} &
  \multicolumn{1}{l|}{18.78/0.93} &
  \multicolumn{1}{l|}{-} &
  \multicolumn{1}{l|}{16.56/0.76} &
  \multicolumn{1}{l|}{25.73/0.93} &
  \multicolumn{1}{l|}{18.48/0.90} &
  \multicolumn{1}{l|}{-} \\ \hline
\multicolumn{1}{|l|}{\nice} &
  \multicolumn{1}{l|}{22.48/0.85} &
  \multicolumn{1}{l|}{23.16/0.90} &
  \multicolumn{1}{l|}{14.34/0.88} &
  \multicolumn{1}{l|}{18.89/0.80} &
  \multicolumn{1}{l|}{26.38/0.91} &
  \multicolumn{1}{l|}{18.11/0.93} &
  \multicolumn{1}{l|}{-} &
  \multicolumn{1}{l|}{15.45/0.78} &
  \multicolumn{1}{l|}{23.20/0.91} &
  \multicolumn{1}{l|}{17.90/0.92} &
  \multicolumn{1}{l|}{-} \\ \hline
\multicolumn{1}{|l|}{Ours} &
  \multicolumn{1}{l|}{{\color[HTML]{3531FF} 24.38/0.86}} &
  \multicolumn{1}{l|}{{\color[HTML]{3531FF} 24.91/0.90}} &
  \multicolumn{1}{l|}{{\color[HTML]{3531FF} 19.31/0.93}} &
  \multicolumn{1}{l|}{{\color[HTML]{3531FF} 22.77/0.82}} &
  \multicolumn{1}{l|}{{\color[HTML]{3531FF} 27.01/0.92}} &
  \multicolumn{1}{l|}{{\color[HTML]{3531FF} 20.71/0.95}} &
  \multicolumn{1}{l|}{16.75/0.88} &
  \multicolumn{1}{l|}{{\color[HTML]{3531FF} 20.04/0.80}} &
  \multicolumn{1}{l|}{{\color[HTML]{3531FF} 27.15/0.93}} &
  \multicolumn{1}{l|}{{\color[HTML]{3531FF} 20.97/0.95}} &
  \multicolumn{1}{l|}{15.23/0.84} \\ \hline
\multicolumn{12}{|c|}{\begin{tabular}[c]{@{}c@{}}Depth Reconstruction (PSNR $\uparrow$ / L1 $\downarrow$)\xrowht{15pt}\end{tabular}} \\ \hline
\multicolumn{1}{|c|}{} &
  \multicolumn{3}{c|}{SYN-Scene-A} &
  \multicolumn{4}{c|}{SYN-Scene-B} &
  \multicolumn{4}{c|}{SYN-Scene-C} \\ \hline
\multicolumn{1}{|c|}{} &
  \multicolumn{1}{c|}{Full} &
  \multicolumn{1}{c|}{BG} &
  \multicolumn{1}{c|}{FG} &
  \multicolumn{1}{c|}{Full} &
  \multicolumn{1}{c|}{BG} &
  \multicolumn{1}{c|}{FG} &
  \multicolumn{1}{c|}{FG2} &
  \multicolumn{1}{c|}{Full} &
  \multicolumn{1}{c|}{BG} &
  \multicolumn{1}{c|}{FG} &
  FG2 \\ \hline
\multicolumn{1}{|l|}{\imap} &
  \multicolumn{1}{l|}{17.87/0.67} &
  \multicolumn{1}{l|}{22.96/0.45} &
  \multicolumn{1}{l|}{14.99/0.27} &
  \multicolumn{1}{l|}{16.49/0.96} &
  \multicolumn{1}{l|}{20.23/0.66} &
  \multicolumn{1}{l|}{16.74/0.10} &
  \multicolumn{1}{l|}{-} &
  \multicolumn{1}{l|}{16.43/1.63} &
  \multicolumn{1}{l|}{21.22/1.27} &
  \multicolumn{1}{l|}{16.74/0.19} &
  \multicolumn{1}{l|}{-} \\ \hline
\multicolumn{1}{|l|}{\nice} &
  \multicolumn{1}{l|}{16.23/0.62} &
  \multicolumn{1}{l|}{20.84/0.38} &
  \multicolumn{1}{l|}{13.90/0.32} &
  \multicolumn{1}{l|}{12.64/1.16} &
  \multicolumn{1}{l|}{14.82/0.79} &
  \multicolumn{1}{l|}{16.03/0.12} &
  \multicolumn{1}{l|}{-} &
  \multicolumn{1}{l|}{17.50/0.84} &
  \multicolumn{1}{l|}{26.32/0.39} &
  \multicolumn{1}{l|}{16.68/0.19} &
  \multicolumn{1}{l|}{-} \\ \hline
\multicolumn{1}{|l|}{Ours} &
  \multicolumn{1}{l|}{{\color[HTML]{3531FF} 23.19/0.26}} &
  \multicolumn{1}{l|}{{\color[HTML]{3531FF} 26.70/0.18}} &
  \multicolumn{1}{l|}{{\color[HTML]{3531FF} 18.59/0.11}} &
  \multicolumn{1}{l|}{{\color[HTML]{3531FF} 21.81/0.33}} &
  \multicolumn{1}{l|}{{\color[HTML]{3531FF} 26.67/0.20}} &
  \multicolumn{1}{l|}{{\color[HTML]{3531FF} 18.75/0.07}} &
  \multicolumn{1}{l|}{14.96/0.19} &
  \multicolumn{1}{l|}{{\color[HTML]{3531FF} 23.60/0.45}} &
  \multicolumn{1}{l|}{{\color[HTML]{3531FF} 31.28/0.27}} &
  \multicolumn{1}{l|}{{\color[HTML]{3531FF} 19.78/0.11}} &
  \multicolumn{1}{l|}{13.73/0.50} \\ \hline
\end{tabular}%
}

\end{table*}
\begin{table}[]
\centering
\mycaption{Reconstruction error on \behave}{
We report total scene reconstruction errors using the training camera k0 due to the lack of validation views and per-frame annotation.
Our method consistently produces better reconstruction quality benefiting from the proposed joint optimization and the deformation module. See Figure~\ref{fig:comparison_real} for qualitative evaluation.
}
\label{tab_eval_bh_train}
\resizebox{\columnwidth}{!}{%
\begin{tabular}{|ccccc|}
\hline
\multicolumn{5}{|c|}{\begin{tabular}[c]{@{}c@{}}Color Reconstruction (PSNR $\uparrow$ / SSIM $\uparrow$)\xrowht{15pt}\end{tabular}} \\ \hline
\multicolumn{1}{|c|}{} &
  \multicolumn{1}{c|}{tablesquare\_move} &
  \multicolumn{1}{c|}{trashbin} &
  \multicolumn{1}{c|}{yogaball\_play} &
  chairblack\_lift \\ \hline
\multicolumn{1}{|c|}{\imap} &
  \multicolumn{1}{c|}{14.68 / 0.66} &
  \multicolumn{1}{c|}{13.46 / 0.65} &
  \multicolumn{1}{c|}{12.42 / 0.64} &
  13.28 / 0.64 \\ \hline
\multicolumn{1}{|c|}{\nice} &
  \multicolumn{1}{c|}{11.35 / 0.54} &
  \multicolumn{1}{c|}{11.71 / 0.55} &
  \multicolumn{1}{c|}{12.16 / 0.60} &
  13.30 / 0.63 \\ \hline
\multicolumn{1}{|c|}{Ours} &
  \multicolumn{1}{c|}{{\color[HTML]{3531FF} 26.48 / 0.85}} &
  \multicolumn{1}{c|}{{\color[HTML]{3531FF} 27.75 / 0.87}} &
  \multicolumn{1}{c|}{{\color[HTML]{3531FF} 28.03 / 0.87}} &
  {\color[HTML]{3531FF} 26.71 / 0.85} \\ \hline
\end{tabular}%
}
\resizebox{\columnwidth}{!}{%
\begin{tabular}{|ccccc|}
\hline
\multicolumn{5}{|c|}{\begin{tabular}[c]{@{}c@{}}Depth Reconstruction (PSNR $\uparrow$ / L1 $\downarrow$) \xrowht{15pt}\end{tabular}} \\ \hline
\multicolumn{1}{|c|}{} &
  \multicolumn{1}{c|}{tablesquare\_move} &
  \multicolumn{1}{c|}{trashbin} &
  \multicolumn{1}{c|}{yogaball\_play} &
  chairblack\_lift \\ \hline
\multicolumn{1}{|c|}{\imap} &
  \multicolumn{1}{c|}{24.84 / 0.31} &
  \multicolumn{1}{c|}{21.67 / 0.49} &
  \multicolumn{1}{c|}{22.25 / 0.43} &
  26.21 / 0.28 \\ \hline
\multicolumn{1}{|c|}{\nice} &
  \multicolumn{1}{c|}{25.48 / 0.23} &
  \multicolumn{1}{c|}{21.89 / 0.44} &
  \multicolumn{1}{c|}{22.62 / 0.38} &
  27.00 / 0.24 \\ \hline
\multicolumn{1}{|c|}{Ours} &
  \multicolumn{1}{c|}{{\color[HTML]{3531FF} 30.06 / 0.14}} &
  \multicolumn{1}{c|}{{\color[HTML]{3531FF} 30.07 / 0.15}} &
  \multicolumn{1}{c|}{{\color[HTML]{3531FF} 30.29 / 0.13}} &
  {\color[HTML]{3531FF} 30.39 / 0.14} \\ \hline
\end{tabular}%
}
\end{table}

\begin{figure}[b!]
    \centering 
    \includegraphics[width=0.95\columnwidth]{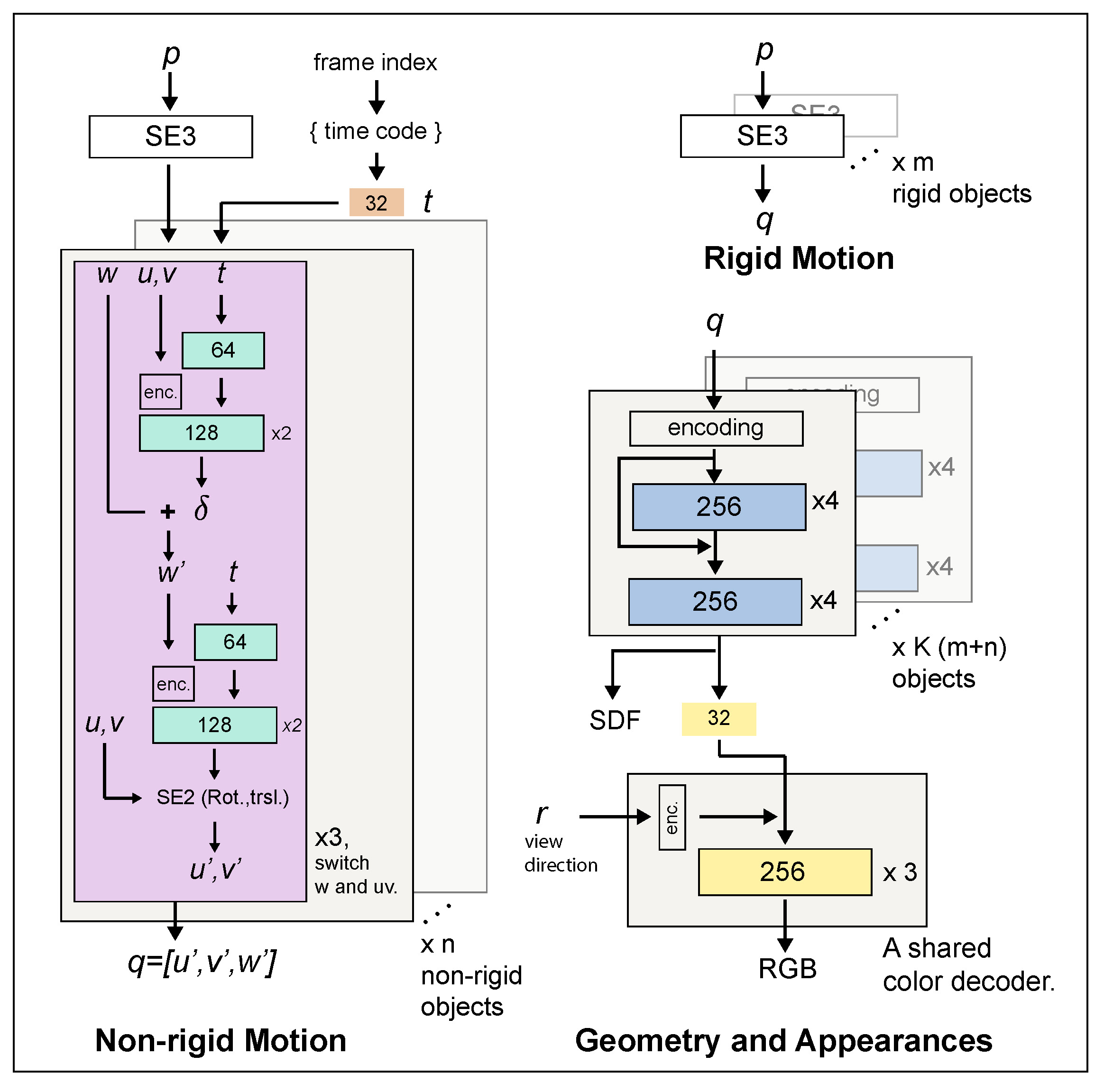}
    \mycaption{Our network architecture}{We use sine positional encoding as NeRF~\cite{martin2021nerf}. The number of rigid and non-rigid motion blocks depends on the objects' motion labels. 
    We employ a bijective deformation block~\cite{cai2022neural} for each non-rigid object.
    Unlike \cite{park2021hypernerf,cai2022neural}, we do not predict ambient coordinates in the non-rigid motion block.}
    \label{fig:arch}
\end{figure}

\para{Architecture.} Figure~\ref{fig:arch} shows our network architecture. 
For the geometry network, we use an SDF field with geometric initialization~\cite{gropp2020implicit}, weighted normalization~\cite{salimans2016weight}, Softplus activations, and a skip-connection MLP. The input coordinates and view directions are lifted to a high dimensional space using positional encoding~\cite{martin2021nerf}. For rigid objects, we use $\mathbb{SE}3$ representation, i.e., a quaternion and a translation vector. For non-rigid objects, we use bijective deformation blocks~\cite{cai2022neural} with Softplus activations. For the color MLP network, we use ReLU activation. 

\para{Model size and implementation details.}
We report the model size of our methods and comparisons.
\imap uses 0.9MB~(FG/BG); 
\nice uses 76MB~(FG) and 135MB~(BG) with $32^3$+$64^3$ grid resolutions for foreground and $32^3$+$80^3$ for background. In contrast, our model takes 5.7MB for the whole scene.
We train all methods using our training framework on a single Nvidia RTX 3090 GPU. We do not use input depth to guide ray sampling for any of the methods as we observed that this reduces models' generalization ability. Instead, at each training iteration, we perform inverse transform sampling and sample $256$ rays with $128$ points per ray.

\para{Comparison.}
We compare our approach against different competing alternatives. 
Existing monocular approaches can be categorized as either employing an MLP (e.g., \imap~\cite{sucar2021imap}),  or using multi-resolution feature grids (e.g., \nice~\cite{zhu2022nice}).
Since these competing methods do not support joint optimizing multiple objects, we \textit{additionally provide} the segmentation and poses generated from our initialization step (see Section~\ref{sec:algorithm}) and \textit{manually} run them multiple times to reconstruct background and dynamic objects.
Note that we modified the ray sample step of \imap and \nice to accept object segmentation input, and we use L1 segmentation loss when training foreground models. 
For both \imap and \nice, we employ the open-source network implementation~\cite{zhu2022nice} in our training framework instead of their multi-threads SLAM framework, which contains several optimizations (e.g., view-purging) for real-time applications.
Note that \imap, with provided background and object segmentation information, can be seen as an upper bound for performance of a method like RigidFusion~\cite{wong2021rigidfusion}. 

\begin{figure*}[!]
    \centering
    \includegraphics[width=0.85\textwidth]{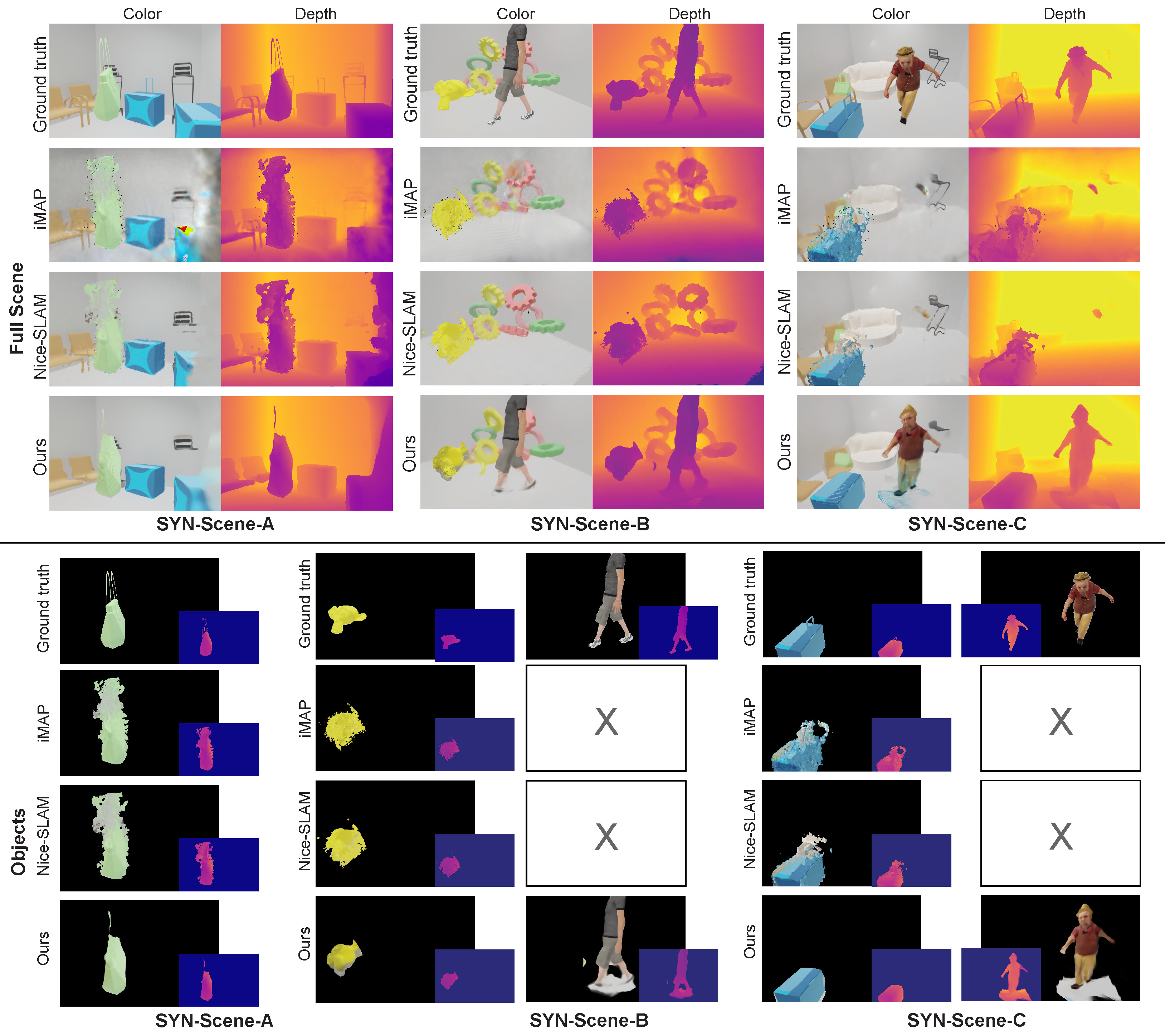} 
    \mycaption{Comparisons of scene and object reconstruction on our synthetic dataset}{
    Visually comparing ours against \imap~\cite{sucar2021imap} and \nice~\cite{zhu2022nice} on our synthetic sequences using the validation cameras. See Table~\ref{tab_eval_syn_novel} for quantitative evaluation. 
    Note that the other methods fail to produce any reconstruction for the nonrigidly moving human. 
    Further, our results are higher in quality and capture finer geometric (e.g., the handle of the green bag) and appearance details (e.g., shading on the yellow monkey face). 
    Please note that Scene C shows a challenging validation frame where the human undergoes a strong deformation. Handling it requires more regularization to force the network to learn the non-rigid motion.
    }
    \label{fig:comparison_syn}
\end{figure*}

\begin{figure*}[!]
    \centering 
    \includegraphics[width=0.9\textwidth]{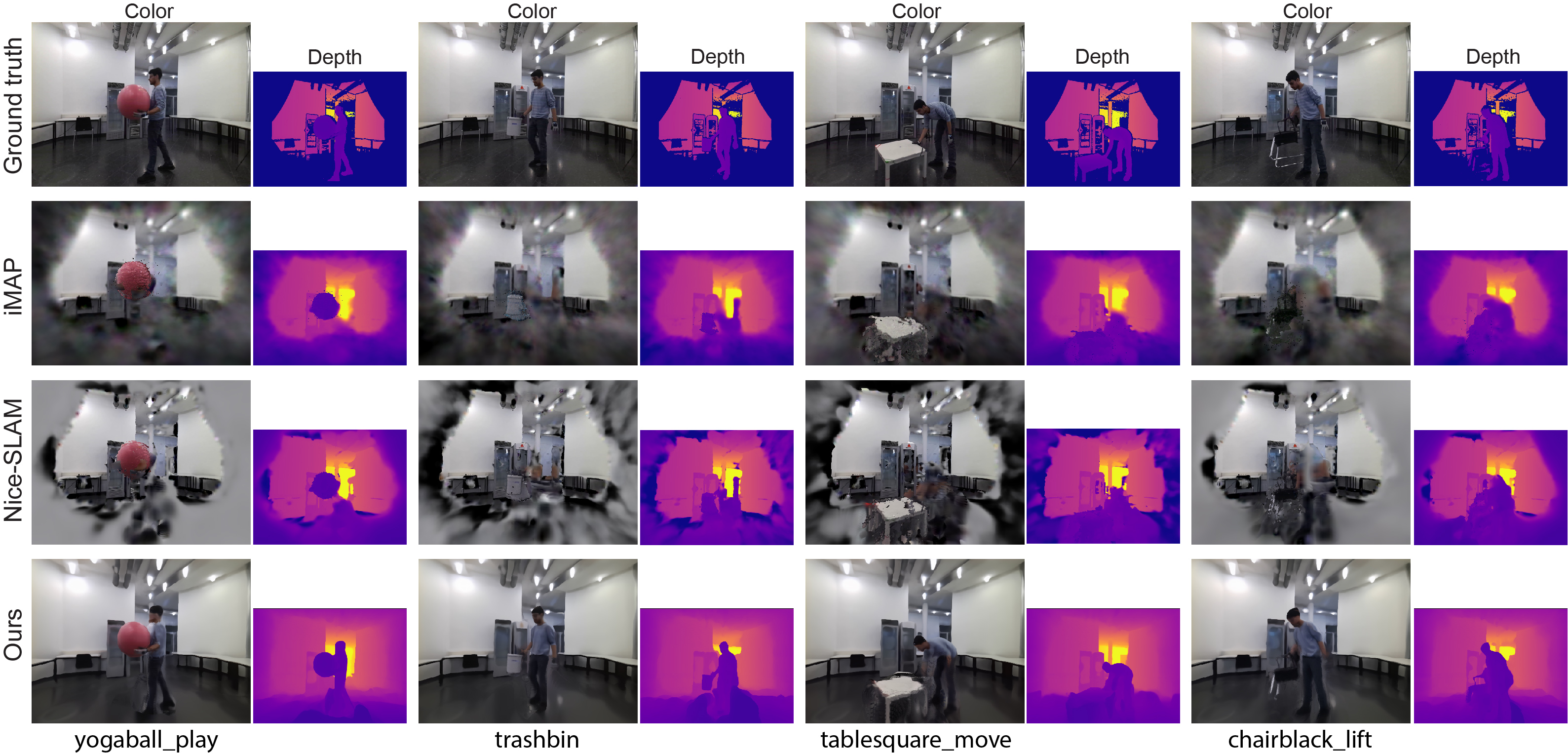}   
    \mycaption{Comparisons of scene reconstruction on the \behave dataset}{
    Visually comparing our results against \imap~\cite{sucar2021imap} and \nice~\cite{zhu2022nice} on the \behave sequences. 
    Notably, our method generalizes better when the scene contains large missing depth areas, showing the learned geometry model is constrained well (see the wall in the training views).
    See Table~\ref{tab_eval_bh_train} for quantitative evaluation.
    }
    \label{fig:comparison_real}
\end{figure*} 

\begin{figure*}[!]
    \centering
    \begin{overpic}[width=0.96\textwidth]{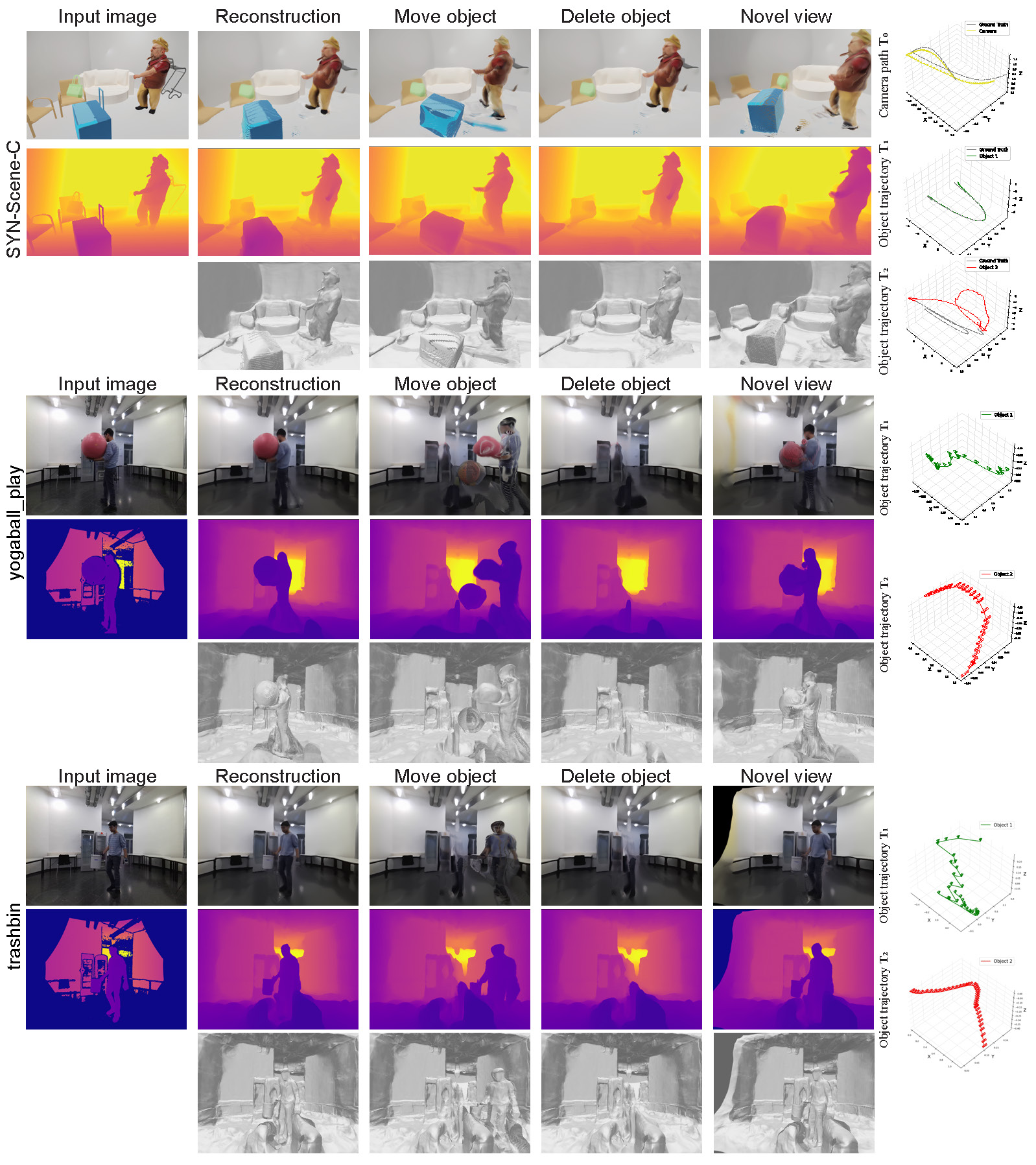}
    \end{overpic}
    \mycaption{Reconstruction and Applications}{
    Reconstruction quality and enabled applications on two synthetic scenes. Please refer to the supplemental videos. Here we show frames for the output RGB, depth, and underlying recovered geometries (extracted by running Marching cubes on the estimated implicit representations $\pSDF^i$). 
    We also show the recovered trajectories, along with corresponding ground truth trajectories. Recall that the $0$-th object being the background, and $\{\transform{0}(t)\}$ represents the camera path. Any stationary object gets reconstructed in the background layer in our factorization. 
    We observed some artifacts caused by unseen geometry (e.g., move objects examples in the second and third rows) and ambiguous decomposition (e.g., the blue box in the first row), because we only have access to monocular and partially occluded input.
    }
    \label{fig:results_plate}
\end{figure*}

\para{Evaluation metrics.} 
We compare different methods across a range of metrics. 
We evaluate novel view \textit{rendering quality} using PSNR, SSIM, and L1 for reconstruction quality in Table~\ref{tab_eval_syn_novel} and Table~\ref{tab_eval_bh_train}.
We also qualitatively evaluate resynthesis quality under the authoring of updated object trajectory as well for addition or deletion of objects from the factored scenes in Figure~\ref{fig:results_plate}. 

\para{Qualitative evaluation.} 
In Figure~\ref{fig:comparison_syn},~\ref{fig:comparison_real}, and~\ref{fig:results_plate}, we qualitatively compare our method against alternative approaches (\imap and \nice). 
Note that although the comparison approaches jointly learn for scene geometry and appearance, they assume the scenes to be static. 
In other words, these methods provide only partial factorization into scene models and camera trajectories.
Thus, we run them multiple times with the same segmentation and pose initialization to reconstruct background and dynamic objects.
Please check the supplemental webpage for result comparisons. 

Our method produces better quality on both synthetic and real-world scans,  both appearance and geometry.
Figure~\ref{fig:comparison_syn} demonstrates our joint optimization scheme improves the object segmentation leading to clearer geometry. 
Figure~\ref{fig:comparison_real} shows the comparison of scene reconstruction on the \behave sequences with large missing depth areas. Our method generalizes better than the comparison.
In Figure~\ref{fig:results_plate}, we also present the extracted object motion trajectories in $\mathbb{R}^3$ as recovered by our initialization step. For the synthetic example, we add groundtruth trajectories (gray colored) for comparison. Note that since we do not perform any loop closure, the trajectory estimates degrade over a longer distance due to error accumulation.

\para{Quantitative evaluation.}
We present a quantitative comparison in Table~\ref{tab_eval_syn_novel} for reconstruction quality using the validation cameras, separately for RGB and depth channels. 
Notably, our method consistently outputs better reconstruction than others (\imap and \nice), indicating that our sampling scheme extracts a proper factorization and hence avoids overfitting to training views. 
In the absence of ground truth and validation views, we cannot run quantitative evaluation for real sequences (Figure~\ref{fig:comparison_real}).

\begin{figure}[!]
\vspace{-10pt}
\centering
\includegraphics[width=0.35\textwidth]{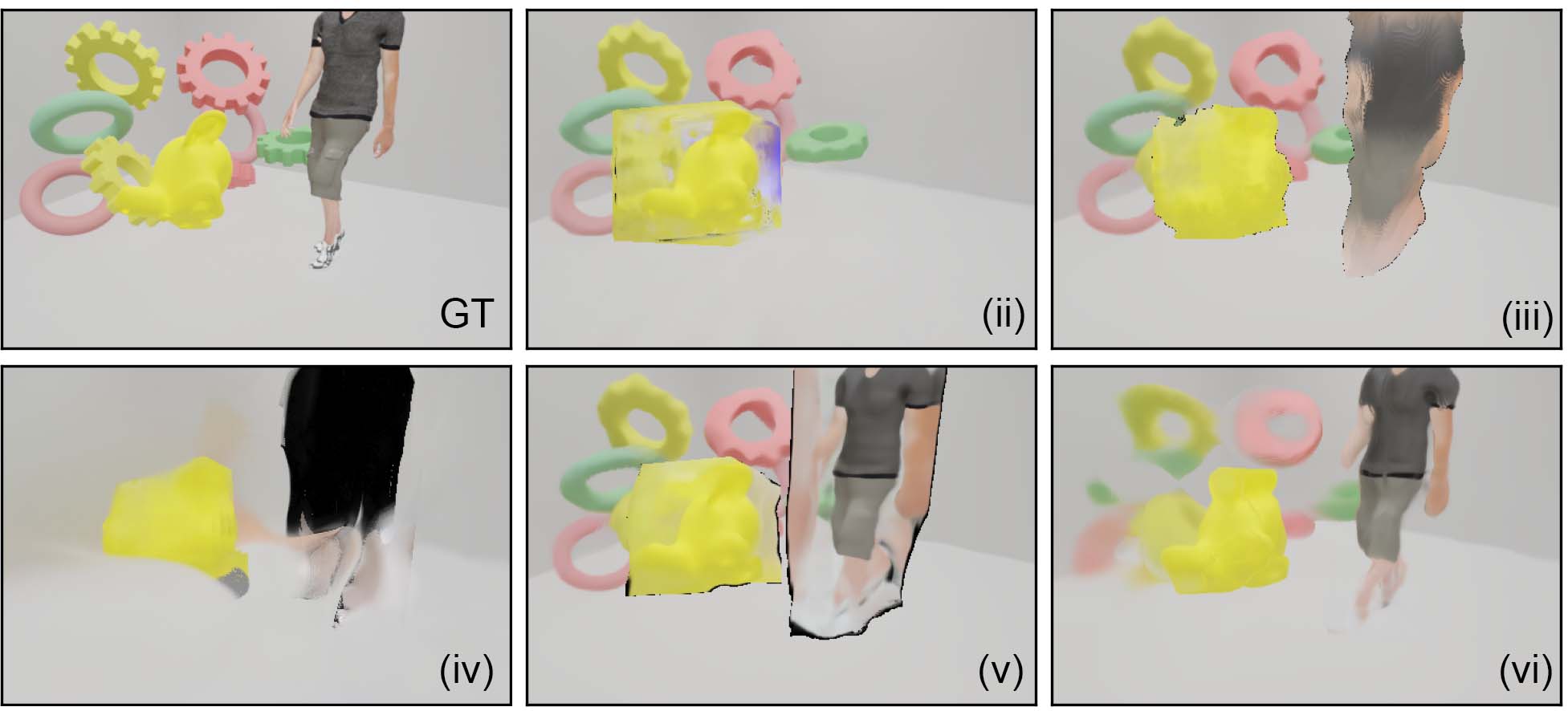}
\mycaption{Ablation study and reconstruction results}{
The number indicates the setting in Table~\ref{tab_ablation}. 
Surface regularizers (iii) enforce the network to learn geometry. The segmentation loss in joint setting (iv) performed poorly due to the conflicted signals between each network, which may require per-object rendering during training. 
}
\label{fig_ablation}
\vspace{-10pt}
\end{figure}

\para{Ablation study.}
In Table~\ref{tab_ablation} and Figure~\ref{fig_ablation}, we conduct an ablation study using our synthetic dataset.
While the commonly employed segmentation loss \cite{wang2021neus,yariv2021volume,cai2022neural} can constrain the object shape through the rendered mask (weights of each sampled ray), it blocks the foreground reconstruction in joint optimization. 
The surface regularizers can stabilize the geometry models and improve both color and depth reconstruction.   
Our final setting (with surface regularizers and freespace loss) has the best full-scene reconstruction quality. 
The segmentation loss fights with thee reconstruction loss in the joint training setting, and the implicit networks fail to learn object surface. Therefore, we replace the segmentation loss with freespace loss allowing the network to optimize all objects and learn correct object geometry.
\begin{table}[]
\vspace{-10pt}
\centering 
\mycaption{Ablation study on our synthetic dataset}{
We evaluate total scene reconstruction errors using the validation cameras on our synthetic dataset.
Segment. Loss: supervise the rendered masks (weights of each sampled ray) using the input segmentation~\cite{wang2021neus,yariv2021volume,cai2022neural}. 
Recon. Loss: color and depth reconstruction loss. 
Surface Reg. and Freespace Loss: the surface regularizer and the loss described in Section~\ref{sec:algorithm}.
We observed that the commonly employed segmentation loss is unsuitable for supervising multiple implicit networks (iv) and causes a large performance drop. Also, the freespace loss is not performed well in single object setting (v) due to the lack of background information. See Figure~\ref{fig_ablation} for the visualization. Our setting (vi) performs well and is able to handle imperfect segmentation input.}

\label{tab_ablation}
\resizebox{\columnwidth}{!}{%
\begin{tabular}{|cccccc|cc|}
\hline
\multicolumn{6}{|c|}{Ablation Settings\xrowht{15pt}} &
  \multicolumn{2}{c|}{Scene Reconstruction\xrowht{15pt}} 
  \\ \hline
  \multicolumn{1}{|l|}{} &
  \multicolumn{1}{l|}{\begin{tabular}[c]{@{}l@{}}Recon. \\ Loss\end{tabular}} &
  \multicolumn{1}{l|}{\begin{tabular}[c]{@{}l@{}}Segment. \\ Loss\end{tabular}} &
  \multicolumn{1}{l|}{\begin{tabular}[c]{@{}l@{}}Surface \\ Reg.\end{tabular}} &
  \multicolumn{1}{l|}{\begin{tabular}[c]{@{}l@{}}Freespace \\ Loss\end{tabular}} &
  \multicolumn{1}{l|}{\begin{tabular}[c]{@{}l@{}}Joint \\ Training\end{tabular}} &
  \multicolumn{1}{c|}{\begin{tabular}[c]{@{}c@{}}Color\\ (PSNR$\uparrow$/ SSIM$\uparrow$)\end{tabular}} &
  \multicolumn{1}{c|}{\begin{tabular}[c]{@{}c@{}}Depth\\ (PSNR$\uparrow$/ L1$\downarrow$)\end{tabular}} 
  \\ \hline
  \multicolumn{1}{|c|}{i} &
  \multicolumn{1}{c|}{\vv} &
  \multicolumn{1}{c|}{} &
  \multicolumn{1}{c|}{} &
  \multicolumn{1}{c|}{} &
   &
  \multicolumn{1}{c|}{18.59 / 0.81} & 
  \multicolumn{1}{c|}{16.54 / 0.65} 
  \\ \hline
  \multicolumn{1}{|c|}{ii} &
  \multicolumn{1}{c|}{\vv} &
  \multicolumn{1}{c|}{\vv} &
  \multicolumn{1}{c|}{} &
  \multicolumn{1}{c|}{} &
   &
  \multicolumn{1}{c|}{18.87 / 0.82} & 
  \multicolumn{1}{c|}{17.31 / 0.57} 
   \\ \hline
  \multicolumn{1}{|c|}{iii} &
  \multicolumn{1}{c|}{\vv} &
  \multicolumn{1}{c|}{\vv} &
  \multicolumn{1}{c|}{\vv} &
  \multicolumn{1}{c|}{} &
   &
  \multicolumn{1}{c|}{20.08 / 0.79} & 
  \multicolumn{1}{c|}{18.34 / 0.49} 
   \\ \hline
  \multicolumn{1}{|c|}{iv} &
  \multicolumn{1}{c|}{\vv} &
  \multicolumn{1}{c|}{\vv} &
  \multicolumn{1}{c|}{\vv} &
  \multicolumn{1}{c|}{} &
  \vv &
  \multicolumn{1}{c|}{14.47 / 0.74} & 
  \multicolumn{1}{c|}{\;8.48 / 3.07} 
  \\ \hline
  \multicolumn{1}{|c|}{v} &
 \multicolumn{1}{c|}{\vv} &
  \multicolumn{1}{c|}{} &
  \multicolumn{1}{c|}{\vv} &
  \multicolumn{1}{c|}{\vv} &
   &
  \multicolumn{1}{c|}{17.85 / 0.76} & 
  \multicolumn{1}{c|}{14.85 / 0.85} 
  \\ \hline
  \multicolumn{1}{|c|}{vi} &
  \multicolumn{1}{c|}{\vv} &
  \multicolumn{1}{c|}{} &
  \multicolumn{1}{c|}{\vv} &
  \multicolumn{1}{c|}{\vv} &
  \vv &
  \multicolumn{1}{c|}{{\color[HTML]{0000FF} 22.78 / 0.84}} & 
  \multicolumn{1}{c|}{{\color[HTML]{0000FF} 20.99 / 0.42}}  
  \\ \hline
\end{tabular}%
}
\vspace{-8pt}
\end{table}

\begin{figure*}[t!]
    \centering 
    \begin{overpic}[width=0.85\textwidth]{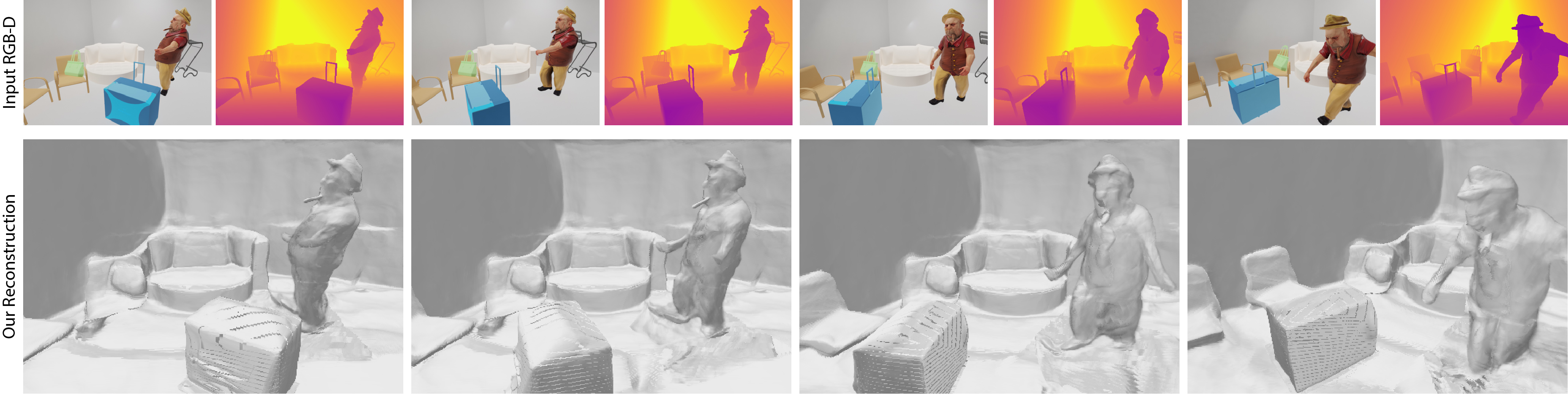}
    \end{overpic}
    \mycaption{Dynamic scene reconstruction}{
    We demonstrate our full scene reconstruction exhibiting non-rigid object deformation across different time indexes. Note how ours can recover plausible movement of the movement of the limbs across time. 
    }
    \label{fig:rect_example}
\end{figure*} 
\para{Model size and reconstruction quality.}
We conduct another ablation study to examine the effect of model size using our synthetic dataset. We adjust the hidden dimension size and set up three models: small, medium, and big, with 370K, 1381K, 5342K parameters, respectively. We trained all models for 100K iterations and observed that the reconstruction quality increased linearly when more parameters were used. The result color PSNR values are 22.9 (small), 23.1 (medium), and 23.18 (big); and the depth L1 errors are 0.35 (small), 0.33 (medium), and 0.32 (big).

\para{Applications.}
We demonstrate three different editing modes in Figure~\ref{fig:results_plate}: (i)~novel view synthesis by changing the extracted camera trajectory; (ii)~object level manipulation by changing one or more object trajectories; (iii)~deleting objects by removing them from the factored representations. Note that the scene-specific learned renders are held fixed during any of the edits. 
While we only train with monocular input, our model can still support editing and output reasonable reconstruction.
These edit modes are be applied separately or in parallel, and test the quality of the scene understanding (i.e., factorization) by revealing unseen object parts and configurations. 
These editing operations are non-trivial because our model is supervised using monocular input containing large motion. 
Removing the artifacts in Figure~\ref{fig:results_plate} will be interesting future work.

\section{Conclusion}

We have presented \textit{factored neural representation} along with a joint optimization formulation that allows  to separate a monocular RGB-D video into object level encodings, without requiring access to additional shape or motion priors. We demonstrated how to  directly obtain object level coupled geometry and appearance encoding, along with object trajectories and deformations. The factorized representation directly supports novel view synthesis along with authoring edits on object trajectories. 
Our work has limitations that we want to address in future works, as discussed next. 

\para{Joint camera and object tracking.} In our current implementation, we do not optimize the camera obtained during the initialization phase. It  would be interesting to jointly finetune the initial estimates, possibly by loop closing and locally linearizing the transformation estimates to simplify the resultant optimization.

\para{Inter object interactions and shading.} In this paper, we do not model object-object or object-background effects. For example, we do not explicitly model shadows~\cite{wu2022d}, reflections~\cite{guo2022nerfren}, transparency~\cite{IchnowskiAvigal2021DexNeRF}, or object interactions arising from human affordance considerations. In the future, it would be a possibility to model these in the volume rendering step.

\para{Better architecture.} At present, we modeled object functions of the form $f_\theta$ simply using MLPs. More recent alternatives and localized versions like hashing~\cite{muller2022instant} or direct functions~(e.g., ReluFields~\cite{karnewar2022relu}) can be alternatively explored. However, the challenge would then be to effectively integrate information across multiple frames to model deformations, possibly by dynamically reindexing the local grid-based representations. 

\para{Shape priors.} As our method does not rely on any object or motion priors, it cannot recover from significant occlusions. We plan to regularize the problem by incorporating data priors, and possibly reducing the dimensions of the variables by working in a learned latent space. However, even deciding which representation to use to anchor such a learned shape space for arbitrary objects still remains an open research topic. 

\para{Acknowledgement}
This work was supported by the Marie Skłodowska-Curie grant number 956585, the UCL AI Centre, and gifts from Adobe.

\bibliographystyle{eg-alpha-doi}

\bibliography{source/dynamicNerf_bib}

\end{document}